\documentclass{article}

\usepackage{microtype}
\usepackage{graphicx}
\usepackage{subfigure}
\usepackage{booktabs} 

\usepackage{hyperref}
\usepackage[utf8]{inputenc} 

\usepackage[accepted]{icml2025}

\usepackage{amsmath}
\usepackage{amssymb}
\usepackage{mathtools}
\usepackage{amsthm}

\usepackage[capitalize,noabbrev]{cleveref}

\theoremstyle{plain}

\theoremstyle{definition}

\theoremstyle{remark}

\usepackage[textsize=tiny]{todonotes}

\begin{document}
\icmltitlerunning{Self-supervised Heterogeneous Graph Neural Network with Optimal
Transport}

\twocolumn[
\icmltitle{HGOT: Self-supervised Heterogeneous Graph Neural Network with Optimal Transport}




\begin{icmlauthorlist}
\icmlauthor{Yanbei Liu}{1}
\icmlauthor{Chongxu Wang}{1}
\icmlauthor{Zhitao Xiao}{1}
\icmlauthor{Lei Geng}{1}
\icmlauthor{Yanwei Pang}{2}
\icmlauthor{Xiao Wang}{3}
\end{icmlauthorlist}

\icmlaffiliation{1}{School of Life Sciences, Tiangong University, Tianjin, China}
\icmlaffiliation{2}{School of Electrical and Infomation Engineering, Tianjin University, Tianjin, China}
\icmlaffiliation{3}{School of Software, Beihang University, Beijing, China}

\icmlcorrespondingauthor{Zhitao Xiao}{xiaozhitao@tiangong.edu}
\icmlcorrespondingauthor{Xiao Wang}{ xiao\_wang@buaa.edu }

\icmlkeywords{Machine Learning, ICML}

\vskip 0.3in
]



\printAffiliationsAndNotice{}  

\begin{abstract}
Heterogeneous Graph Neural Networks (HGNNs), have demonstrated excellent capabilities in processing heterogeneous information networks. Self-supervised learning on heterogeneous graphs, especially contrastive self-supervised strategy, shows great potential when there are no labels. However, this approach requires the use of carefully designed graph augmentation strategies and the selection of positive and negative samples. 
Determining the exact level of similarity between sample pairs is non-trivial.
To solve this problem, we propose a novel self-supervised Heterogeneous graph neural network with Optimal Transport (HGOT) method which is designed to facilitate self-supervised learning for heterogeneous graphs without graph augmentation strategies. Different from traditional contrastive self-supervised learning, HGOT employs the optimal transport mechanism to relieve the laborious sampling process of positive and negative samples. Specifically, we design an aggregating view (central view) to integrate the semantic information contained in the views represented by different meta-paths (branch views). Then, we introduce an optimal transport plan to identify the transport relationship between the semantics contained in the branch view and the central view. This allows the optimal transport plan between graphs to align with the representations, forcing the encoder to learn node representations that are more similar to the graph space and of higher quality.
Extensive experiments on four real-world datasets demonstrate that our proposed HGOT model can achieve state-of-the-art performance on various downstream tasks. In particular, in the node classification task, HGOT achieves an average of more than 6\% improvement in accuracy compared with state-of-the-art methods. 
\end{abstract}

\section{Introduction}

Heterogeneous graphs are prevalent in numerous real-world contexts, including academic networks (\textcolor[RGB]{25,25,112}{Wang et al. 2019}),  biomedical networks (\textcolor[RGB]{25,25,112}{Bai et al. 2021}),  social networks (\textcolor[RGB]{25,25,112}{Cao et al. 2021}) and so on. These graphs serve to represent diverse relationships among various types of nodes. In recent years, a range of heterogeneous graph neural networks (HGNNs) has been developed to effectively capture the intricate structures and abundant semantic information inherent in these graphs, demonstrating commendable performance in the domain of heterogeneous graph representation learning (\textcolor[RGB]{25,25,112}{Wang et al. 2020},  \textcolor[RGB]{25,25,112}{Dong et al. 2020}). To date, HGNNs have found extensive applications across a variety of fields  (\textcolor[RGB]{25,25,112}{Tian et al. 2022a}, \textcolor[RGB]{25,25,112}{Bansal et al. 2019}, \textcolor[RGB]{25,25,112}{Wang et al. 2021a}).

The majority of current HGNNs rely on supervised or semi-supervised learning methods (\textcolor[RGB]{25,25,112}{Sun et al. 2025}, \textcolor[RGB]{25,25,112}{Ji et al. 2024}), necessitating the guidance of labeled data throughout the model's learning process. However, acquiring node labels from heterogeneous graph data poses significant challenges and incurs high costs in many applications (\textcolor[RGB]{25,25,112}{Hu et al. 2020}). Consequently, in the absence of labeled data, self-supervised learning (SSL) emerges as a viable alternative, as it seeks to derive valuable supervisory signals directly from the input data. 

In the realm of self-supervised learning applied to heterogeneous graphs, contrastive learning has emerged as a prominent technique. This approach involves carefully designed graph augmentation strategies and the selection of positive and negative samples. The objective is to enhance the similarity between positive samples while concurrently reducing the similarity between negative samples. Some works (\textcolor[RGB]{25,25,112}{Zhao et al. 2024}, \textcolor[RGB]{25,25,112}{Zhu et al. 2022}) on heterogeneous graph contrastive learning  propose different graph augmentation strategies to better capture the underlying high-order semantic structure information. In addition, many methods (\textcolor[RGB]{25,25,112}{Wang et al, 2021a}, \textcolor[RGB]{25,25,112}{Zhao et al. 2024}) retain the traditional contrastive learning and design the method of selecting the required positive and negative sample pairs from different views or different structures, and can more effectively obtain high-level factors. 
Although the contrastive self-supervised learning strategy enhances the quality of node representations in HGNNs and facilitates their application in real-world scenarios devoid of label supervision (\textcolor[RGB]{25,25,112}{Xie et al. 2024}, \textcolor[RGB]{25,25,112}{Liu et al. 2024}, \textcolor[RGB]{25,25,112}{Chen et al. 2024}), two key challenges persist in the contrastive self-supervised learning for heterogeneous graphs:
\begin{itemize}
    \item \textbf{C1: More effective self-supervised learning strategy for heterogeneous graphs.} On the one hand, existing contrastive self-supervised methods on heterogeneous graph require graph augmentations. The heterogeneous graph data structure is discrete, so even if the graph is slightly perturbed, the properties of the graph may change, which leads to inconsistent labels between the original and augmented graphs. Therefore, it is challenging to design reasonable augmentations that guarantee the label consistency for heterogeneous graphs. 
    On the other hand, the strategy of contrastive learning is to maximize the similarity for positive pairs of the same anchor and minimize the similarity for those of negative pairs. However, the concept of “maximum or minimum similarity” is difficult to measure since it is vague and lacks a clear indication of how much similarity should be maximized (or minimized) for a given pair of positive (or negative) views.

    \item \textbf{C2: More comprehensive heterogeneous semantic information aggregation.} 
    Heterogeneous graphs typically exhibit complex structural variations, e.g., objects of different types, diverse relationships, and varying degrees of connectivity. Integrating information across these structures while maintaining the integrity of the graph's overall topology (i.e., the interconnections between objects) presents a further difficulty. A simplistic aggregation approach could distort the graph’s structural properties, leading to a loss of context and nuance.

\end{itemize}

To address these challenges, in this paper, we propose HGOT, a novel self-supervised learning strategy for heterogeneous graphs that does not require graph augmentations and positive or negative samples. We exploits the key concept in optimal transport (OT) theory, which aims to identify the matching relationship between graph and representation. In order to eliminate the diversity of node types, we design the node feature transformation to project all nodes into a unified dimensional feature space. To further capture more comprehensive semantics, we employ the attention mechanism to aggregate different meta-paths (branch views) to obtain an aggregated view (central view) which can integrate the semantic information from different meta-paths. We then introduce an optimal transport plan to identify the transport relationship between the semantics contained in each branch view and the central view. To accurately capture the matching relationships in the original graph space, the encoders are forced to learn representations that exhibit consistency with the optimal transport plans between the corresponding meta-path views and aggregated view. In summary, the main contributions of this work are as follows:

\begin{itemize}
    \item To our best knowledge, this is the first attempt to apply optimal transport theory to heterogeneous graph. By utilizing the optimal transport to calculate the transport plan between the meta-path view and aggregated view, it is no longer necessary to perform data augmentation and provide positive and negative samples for heterogeneous graph self-supervised learning.

    \item We propose a novel self-supervised Heterogeneous Graph neural network with Optimal Transport (HGOT) method, which utilize the optimal transport plan between the local semantics (branch view) and the global semantics (central view) of heterogeneous graphs to align the matching relationship between the graph space and the representation space, thereby obtaining higher-quality node representations.

    \item We conduct extensive experiments on four public benchmarks, demonstrating that the proposed HGOT outperforms the state-of-the-art methods. In particular, in the node classification task, HGOT achieves an average of more than 6\% improvement in accuracy compared with state-of-the-art methods. 
\end{itemize}

\begin{figure*}

    \centering
    
    \includegraphics[width=0.85\linewidth]{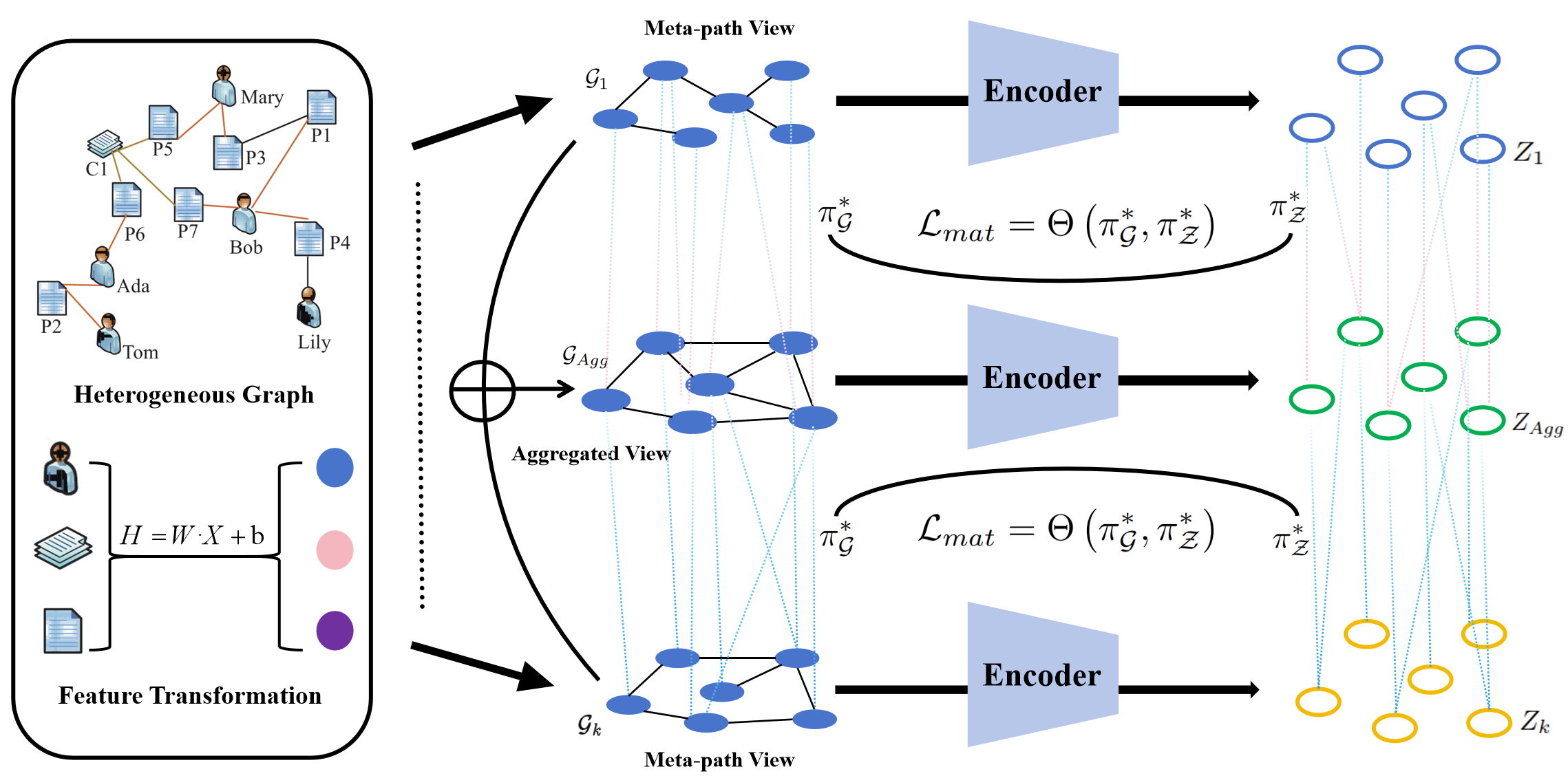}

\caption{\textbf{The overall framework of HGOT:} We first project all nodes' features of the original heterogeneous graph into the same dimensional space. We then design the heterogeneous semantic information aggregation to obtain the meta-path view and aggregated view. Finally, we introduce the optimal transport theory to discover matching relationships between each meta-path view and the aggregated view and align them to obtain higher quality node representations.}
\end{figure*}\textbf{}

\section{Related Work}
\textbf{Heterogeneous graph self-supervised learning.} 
Heterogeneous graph self-supervised learning has been a promising paradigm, and it uses the information of the graph itself to learn node representations. For example,  HeMuc (\textcolor[RGB]{25,25,112}{Zhang et al. 2023}) proposes a contrastive learning method by designing a new sampling strategy that combines structure and characteristic information. HeCo  (\textcolor[RGB]{25,25,112}{Wang et al. 2021a}) applies two network structures to align both local schema and global metapath information. HGCML  (\textcolor[RGB]{25,25,112}{Wang et al. 2023}) constructs a contrastive learning method under multiple views by utilizing meta-paths and proposes a new positive sample sampling strategy to reduce sampling bias. In addition to contrastive learning, generative learning is also a good self-supervised learning method (\textcolor[RGB]{25,25,112}{Kipf and Welling 2016}, \textcolor[RGB]{25,25,112}{Park et al. 2019}). HGMAE (\textcolor[RGB]{25,25,112}{Tian et al. 2023}) applies masked autoencoders to heterogeneous graphs and obtains better representations through the mask mechanism.

\textbf{Optimal Transport.} Optimal transport (OT) (\textcolor[RGB]{25,25,112}{Cédric Villani et al. 2009}) is a mathematical tool widely used in computer vision (\textcolor[RGB]{25,25,112}{Nicolas Bonneel et al. 2011}), generative adversarial network (\textcolor[RGB]{25,25,112}{Cao et al. 2019}) and domain adaptation (\textcolor[RGB]{25,25,112}{Gu et al. 2022}) to calculate the distance between two probability distributions or to align two distributions in different spaces. OT aims to derive a transport plan between a source and a target distribution, such that the transport cost is minimized. (\textcolor[RGB]{25,25,112}{Facundo Mémoli et al. 2011}) proposes the Gromov-Wasserstein (GW) distance between metrics defined within each space which has been used as a distance between graphs in several applications (\textcolor[RGB]{25,25,112}{Vayer Titouan et al. 2020}, \textcolor[RGB]{25,25,112}{Cédric Vincent-Cuaz et al. 2022}, \textcolor[RGB]{25,25,112}{Xu et al. 2019}). In this work, we combine the optimal transport theory with heterogeneous graph neural networks to form a new self-supervised learning paradigm.

\section{Preliminaries}

In this section, we define two concepts related to heterogeneous graphs and two problems of optimal transport on graphs as follows.

\textbf{Definition 3.1. Heterogeneous Graph. }A heterogeneous graph is defined as  $\mathcal{G}=(\mathcal{V},\mathcal{E},T_\mathcal{V},T_\mathcal{E},\varphi, \psi)$, which consists of a node set $\mathcal{V}$, a link set $\mathcal{E}$, and  it is associated with a node type mapping function $\varphi: \mathcal{V} \to T_\mathcal{V}$ and a edge type mapping function $\psi: \mathcal{E} \to T_\mathcal{E}$, where $T_\mathcal{V}$ and $T_\mathcal{E}$ denote sets of object and link types, and $\vert T_\mathcal{V}\vert+\vert T_\mathcal{E}\vert>2$. It is a homogeneous graph with the same node and edge types when $\vert T_\mathcal{V}\vert=1$ and $\vert T_\mathcal{E}\vert=1$. For example, in the academic network ACM, three node types of Paper, Author, and Subject and two types of link relationships constitute a heterogeneous graph.

\textbf{Definition 3.2. Meta-paths.} A meta-path defines a composite relation of several edge types, represented a $\mathcal{P}=T_{\mathcal{V}_1}\stackrel{T_{\mathcal{E}_1}} {\longrightarrow} T_{\mathcal{V}_2} \stackrel{T_{\mathcal{E}_2}} {\longrightarrow} \dots\stackrel{T_{\mathcal{E}_r}}{\longrightarrow} T_{\mathcal{V}_ {\left(\mathrm {r} + \mathrm {1}\right)}}$ (abbreviated as $T_{\mathcal{V}_1}T_{\mathcal{V}_2}\dots T_{\mathcal{V}_ {\left(\mathrm {r} + \mathrm{1}\right)}}$), which describes a composite relation $T_{\mathcal{E}}=T_{\mathcal{E}_1} \circ  
 T_{\mathcal{E}_2} \circ \dots \circ  T_{\mathcal{E}_r}$ between node types $T_{\mathcal{V}_ 1}$ and $T_{\mathcal{V}_ {\left(\mathrm {r} + \mathrm {1}\right)}}$, where $\circ$ denotes the composition operator on relations. Because meta-path is the combination of multiple relations, it contains high-order structures.

\textbf{Problem 3.1. Plan and Optimal Transport.} The optimal transport problem was first proposed by French mathematician Monge to seek the most cost-effective transportation solution for transporting the shape of one pile of sand into the shape of another (\textcolor[RGB]{25,25,112}{Wang et al. 2024}). In particular, given two sets of features $\textbf{\textit{X}}_1= \{\textbf{\textit{X}}_1^i\}_{i=1}^n$  and  $\textbf{\textit{X}}_2= \{\textbf{\textit{X}}_2^j\}_{j=1}^m$ , where $n$ and $m$ are the number of features.  $\mu\in\mathbb{R}^n$ and  $\nu\in\mathbb{R}^m$  are the probability distributions of the entities in the two sets, respectively. It studies how to transform distribution $\mu$ into distribution $\nu$ with the minimum total transportation cost (i.e., the optimal transportation distance). This transport plan is called the optimal transport plan $\pi$, where the element of $\pi$ describes the probability of moving mass from one position to another. 
In summary, the goal of optimal transport is to find a transport plan $\pi$ that transforms the distribution of node set $\textbf{\textit{X}}_1$ into the distribution of node set $\textbf{\textit{X}}_2$ with the minimum transport distance. In this work, we focus on the case of discrete distributions on graphs.

\textbf{Problem 3.2. Optimal Transport for Graphs.} On different graphs, we can view node attribute information as different distributions. If the nodes in the graph have no attributes or label information, we can also use the edge structure of the graph to calculate the optimal transportation plan between two graphs. Given two graphs  $\mathcal{G}_1=(\textbf{\textit{A}}_1, \mu)$  and  $\mathcal{G}_2=(\textbf{\textit{A}}_2, \nu)$ , where $\textbf{\textit{A}}_1$ and $\textbf{\textit{A}}_2$  are the adjacency matrix of two graphs. Similarly, we can compute the optimal transport plan $\pi$ between the edge structures (i.e., adjacency matrix) of the graph. The specific relevant formula for optimal transmission on the graph will be given in detail in the next section. 

\section{Method}

In this section, we formally propose HGOT, a novel heterogeneous graph neural network with optimal transport. In particular, HGOT is constructed by three major components: node feature transformation, aggregated view generation, and multi-view optimal transport alignment. Figure 1 illustrates the framework of HGOT.

\subsection{Node Feature Transformation}

In a heterogeneous graph, different node types have features of different dimensions. Even if their feature vectors have the same dimensions, they may not be in the same feature space.  Feature vectors of different dimensions are troublesome when we process them in a unified framework. In addition, considering that nodes are required to have the same dimensions when using the optimal transport calculation later, we project different types of node features into the same latent vector space. 

Therefore, we apply a type specific linear transformation for each type of nodes by projecting feature vectors into the same latent factor space. Specifically, for the $i$-th node with type $\varphi_i$, we design a type-specific mapping matrix $\textbf{\textit{W}}_{\varphi_i}$ to transform its features into a common latent space as follows:
\begin{equation}
     \textbf{h}_i=\textbf{\textit{W}}_{\varphi_i}\cdot{\textbf{x}_i}+\textbf{b}_{\varphi_i},
\end{equation}
where $\textbf{x}_i$ is the original feature vector, $\textbf{h}_i\in\mathbb{R}^{d\times{1}}$ is the projected feature of $i$-th node, and $\textbf{b}_{\varphi_i}$ denotes as vector bias.

The node feature transformation addresses the heterogeneity of a graph that originates from node features that are not in the same feature space. All nodes’ projected features share the same dimension and are in the same feature space, which facilitates the calculation process of the part of the optimal transport module. 

\subsection{Aggregated View Generation}

Since heterogeneous graphs contain rich semantic information which are reflected by various meta-paths, we design the aggregated view generation mechanism to aggregate different meta-path views. In particular, we decompose the original heterogeneous graph into multiple homogeneous meta-specific graphs according to meta-paths. Then we consider aggregating these homogeneous meta-specific graphs from the perspective of nodes and edges. 

For nodes, following most heterogeneous GNNs' approaches, we first generate multiple views and each view corresponding to one meta-path that encodes one aspect of information. Then we leverage an attentive network to compute specific meta-path embedding $  {\tilde{\textbf{h}}}^{{p}}_i$ for node $v_i$ under the ${p}$-th meta-path as
\begin{equation}
    {\tilde{\textbf{h}}}^{{p}}_i= \mathrm{CONCAT}_ {k=1}^K  \quad\Psi\left(\sum_{{v_j}\in\mathcal{N}_{{p}}\left(v_i\right)}\alpha_{ij}^{{p}}{\mathcal{\textbf{\textit{W}}}}^{{p}}{\textbf{h}_i}\right ),
\end{equation}
where CONCAT concatenates K standalone node representations in each attention head, $\mathcal{N}_{{p}}\left(v_i\right)$ defines the neighborhood of $v_i$ that is connected by meta-path, $\mathcal{\textbf{\textit{W}}}^{{p}}\in{\mathbb{R}^{d\times{m}}}$  is a linear transformation matrix for ${p}$-th meta-path , and $\Psi\left(\cdot\right)$ is the activation function, such as $\mathrm{ReLU}\left(\cdot\right)=\mathrm{max}\left(0,\cdot\right)$. The attention coefficient $\alpha_{ij}^{{p}}$ can be computed by a softmax function
\begin{equation}
\alpha_{ij}^{{p}}=\cfrac{\mathrm{exp}\left(\Psi\left({\textbf{a}}^\top_{p}\left[ {\tilde{\textbf{h}}}^{{p}}_i\Vert{\tilde{\textbf{h}}}^{{p}}_j\right]\right)\right)}{\sum_{v_k\in\mathcal{N}_{p}\left(v_i\right)}{}\mathrm{exp}\left(\Psi\left({\textbf{a}}^\top_{p}\left[ {\tilde{\textbf{h}}}^{{p}}_i\Vert{\tilde{\textbf{h}}}^{{p}}_k\right]\right)\right)},
\end{equation}
where ${\textbf{a}}_{p}\in\mathbb{R}^{2d\times{1}}$ is a trainable linear weight vector for the ${p}$-th meta-path. 

Finally, we fuse the node representations in all views into an aggregated representation. We employ another attentive network to obtain the aggregated representation $\textbf{h}_i^{\rm{agg}}$ that combines information from every semantic space by
\begin{equation}
\textbf{h}_i^{\rm{agg}}=\sum_{p=1}^{\vert{\mathcal{P}\vert}}\beta^{p}\cdot{\tilde{\textbf{h}}}^{p}_i.
\end{equation}
The coefficients are given by
\begin{equation}
    \beta^{p}=\cfrac{\mathrm{exp}\left(\omega^{p}\right)}{\sum_{p'\in{\mathcal{P}}}\mathrm{exp}\left(\omega^{p'}\right)},
\end{equation}
\begin{equation}
    \omega^{p}=\cfrac{1}{\vert{\mathcal{V}}\vert}{\sum_{{v_i}\in{\mathcal{V}}}{{\textbf{q}}\top}\cdot{\mathrm{tanh}\left(\textbf{\textit{M}}{\tilde{\textbf{h}}}^{p}_i+{\tilde{\textbf{b}}}\right)}},
\end{equation}
where ${\textbf{q}}\in\mathbb{R}^{d_m}$ denotes type-level attention vector. $\beta^{p}$ can be interpreted as the contribution of the $p$-th meta-path. 
 ${d_m}\in{\mathbb{R}}$ is a hyperparameter, $\textbf{\textit{M}}\in\mathbb{R}^{{d_m}\times{d}}$, ${{\tilde{\textbf{b}}}}\in{\mathbb{R}^{d_m}}$ are the weight matrix and the bias vector, respectively. Finally, we concatenate and arrange the aggregated representations of all nodes and use a matrix $\textbf{\textit{Z}}_{\rm{agg}}$ to represent the representations of all nodes in the aggregated view. 

For edges, we create the meta-path-based adjacency matrix $\textbf{\textit{A}}_p$ for each meta-path $p\in\mathcal{P}$ via meta-path sampling. In order to directly obtain the edge structure information in the fused view, and do not want to introduce repeated edges, we perform a logical OR operation on all adjacency matrices as follows: 
\begin{equation}
    \textbf{\textit{A}}_{\rm{agg}}={\textbf{\textit{A}}_1}\lor{\textbf{\textit{A}}_2}\lor{...}\lor{\textbf{\textit{A}}_{\vert\mathcal{P}\vert}},
\end{equation}
where $\textbf{\textit{A}}_{\rm{agg}}$ denotes  the adjacency matrix of aggregated view. Note that the operation rule here is to perform an OR operation on each element in the adjacency matrix. 
This obviously means that if two nodes are connected in any meta-path view, then there should be an edge between them in the aggregated view. 

In this step, we obtain the node representation information $\textbf{\textit{Z}}_{\rm{agg}}$ and the edge structure information $\textbf{\textit{A}}_{\rm{agg}}$ of the aggregated view $\mathcal{G}_{\rm{agg}}$. 

\subsection{Multi-view Optimal Transport Alignment }

In order to achieve information matching from subgraphs induced by meta-paths to aggregated graph, we use the theory of optimal transport to calculate the optimal transport plan between each meta-path view and the aggregated view. Then we align the optimal transport plan between graph and representation to achieve information matching. 

Our goal is to find the optimal transport plan for the two views, minimizing the transport cost (i.e., optimal transport distance). Therefore, we choose to start with the transport distance to calculate the optimal transport plan. Specifically, we obtain the node feature matrix $\textbf{\textit{H}}_p$ and $\textbf{\textit{H}}_{\rm{agg}}$ in the meta-path view and the aggregated view after projection in the ‘‘node feature transformation’’ step. The formulation of the OT distance is
\begin{align}
     {\mathcal{{D}_\textup{n}}}\left(\textbf{\textit{H}}_p, \textbf{\textit{H}}_{\rm{agg}}\right)=\min_{\pi\in\Pi\left(\mu, \nu\right)}\sum_{i\in[\![n]\!]}\sum_{j\in[\![m]\!]}\mathcal{C}_X\left(\textbf{\textit{H}}_{p}^{i}, \textbf{\textit{H}}_{\rm{agg}}^{j}\right)\cdot\pi_{ij},
\end{align}

abbreviated as
\begin{align}
     {\mathcal{D}_\textup{n}}\left(\textbf{\textit{H}}_\textit{p}, \textbf{\textit{H}}_{\rm{\textit{}agg}}\right)=\min_{\pi\in\Pi\left(\mu, \nu\right)}\langle{\mathcal{F}\left(\textbf{\textit{H}}_\textit{p}, \textbf{\textit{H}}_{\rm{agg}}\right), \pi}\rangle,
\end{align}

where 
\begin{equation}
  \Pi\left(\mu, \nu\right)=\{\pi\in\mathbb{R}^{n\times{m}} \vert{\pi\cdot{\textbf{1}_{m}}=\mu}, \textbf{1}_{n}\cdot\pi=\nu\} 
\end{equation}
denotes all the joint distributions $\pi$ with the marginal distribution $\mu$ and $\nu$. 
$\mathcal{F}{\left(\textbf{\textit{H}}_{p}, \textbf{\textit{H}}_{\rm{agg}}\right)}_{ij}=\mathcal{C}_{X}{\left(\textbf{\textit{H}}_{p}^{i}, \textbf{\textit{H}}_{\rm{agg}}^{j}\right)}$ is the cost of moving $\textbf{\textit{H}}_{p}^{i}$ to $\textbf{\textit{H}}_{\rm{agg}}^{j}$, we can choose the cosine distance between them to calculate. The \textbf{1} denotes a vector that elements are all 1, $[\![n]\!]=\{{1}, {2}, \dots, {n}\}$, and $\langle{\cdot, \cdot}\rangle$ is the inner product operator. The $\pi\in\mathbb{R}^{n\times{m}}$ is called as transport plan. This distance ${\mathcal{D}_\textup{n}}{\left(\cdot , \cdot\right)}$ is also known as the Wasserstein distance, used to calculate the transport distance between node distribution. 

For graph, it is difficult to measure the cost between two nodes on
different graphs without node label (attribute). Even if there is some way to get the cost between nodes, there is no edge information in the above Wasserstein distance formula. Therefore, we need to take the edge information into account as follows:
\begin{equation}
\begin{split}
 {\mathcal{D}_\textup{e}}{\left(\textbf{\textit{A}}_\textit{p}, \textbf{\textit{A}}_{\textup{agg}}\right)}=\min_{\pi\in\Pi{\left(\mu, \nu\right)}}\sum_{\textit{i}, \textit{j}\in[\![n]\!]^2}\sum_{\textit{k}, \textit{l}\in[\![m]\!]^2}\\\mathcal{C}_{A}{\left(\textbf{\textit{A}}_{p}^{ik}, \textbf{\textit{A}}_{\rm{agg}}^{jl}\right)}\cdot\pi_{ij}\pi_{kl},
 \end{split}
\end{equation}

abbreviated as
\begin{align}
     {\mathcal{D}_\textup{e}}{\left(\textbf{\textit{A}}_\textit{p}, \textbf{\textit{A}}_{\textup{agg}}\right)}=\min_{\pi\in\Pi{\left(\mu, \nu\right)}}\langle{\mathcal{E}{\left(\textbf{\textit{A}}_\textit{p}, \textbf{\textit{A}}_{\textup{agg}}\right)}\otimes\pi, \pi}\rangle,
\end{align}

where $\mathcal{E}{\left(\textbf{\textit{A}}_{p}, \textbf{\textit{A}}_{\textup{agg}}\right)}_{ijkl}=\mathcal{C}_{A}{\left(\textbf{\textit{A}}_{p}^{ik}, \textbf{\textit{A}}_{\textup{agg}}^{jl}\right)}$ and it is a 4-D tensor. $[\![n]\!]^2=[\![n]\!]\times[\![n]\!]$. Since the cost function here involves the adjacency matrix, we directly subtract it and then take the absolute value to calculate, which is $\mathcal{C}_{A}{\left(\textbf{\textit{A}}_{p}^{ik}, \textbf{\textit{A}}_{\textup{agg}}^{jl}\right)}=\vert{\textbf{\textit{A}}_{p}^{ik}-\textbf{\textit{A}}_{\textup{agg}}^{jl}}\vert$. This distance ${\mathcal{D}_\textup{e}}{\left(\cdot , \cdot\right)}$ is also known as the Gromov-Wasserstein distance, used to calculate the transport distance between edge distribution. 
Then we consider the complete graph including node features and edge structure $\mathcal{G}_{p}={\left(\textbf{\textit{H}}_{p}, \textbf{\textit{A}}_{p},\mu\right)}$ and $\mathcal{G}_{\rm{agg}}={\left(\textbf{\textit{H}}_{\textup{agg}}, \textbf{\textit{A}}_{\textup{agg}},\nu\right)}$. The fused Gromov-Wasserstein distance between two graphs can be defined as \begin{equation}
\begin{split}
    {\mathcal{D}_\textup{g}}{\left(\mathcal{G}_{p}, \mathcal{G}_{\textup{agg}}\right)}=\min_{\pi\in\Pi{\left(\mu,\nu\right)}}\sigma\sum_{\textit{ij}}\mathcal{C}_{X}{\left(\textbf{\textit{H}}_{\textit{p}}^{i}, \textbf{\textit{H}}_{\textup{agg}}^{j}\right)}\cdot\pi_{ij}^{\mathcal{G}}\\  +{\left(1-\sigma\right)}\sum_{ijkl}\mathcal{C}_{A}{\left(\textbf{\textit{A}}_{p}^{ik}, \textbf{\textit{A}}_{\textup{agg}}^{jl}\right)}\cdot\pi_{ij}^{\mathcal{G}}\pi_{kl}^{\mathcal{G}},
\end{split}
\end{equation}
which is equivalent to
\begin{small}
    \begin{equation}
    \min_{\pi\in\Pi{\left(\mu, \nu\right)}}\langle{\sigma\mathcal{F}{\left(\textbf{\textit{H}}_{p}, \textbf{\textit{H}}_{\textup{agg}}\right)}+{\left(1-\sigma\right)}\mathcal{E}{\left(\textbf{\textit{A}}_{p}, \textbf{\textit{A}}_{\textup{agg}}\right)}\otimes\pi_{\mathcal{G}}, \pi_{\mathcal{G}}\rangle},
\end{equation}
\end{small}

where $\sigma\in[0;1]$ represents the trade-off parameter for adjusting nodes and edges. The formula is a fusion of Equations (8) and (11).

\textbf{Optimal Transport Plan.} For graphs and representation vectors, they are in different spaces and are essentially two different objects. Although it is challenging to compare two different objects in different spaces, the optimal transport plan provides a solution. In this work, we integrate the node features and topological structure (i.e., edges) on the graph and design a new objective function to align the optimal transport plan between the graph space and the representation space. Specifically, the fused optimal transport plan $\pi^*_{\mathcal{G}}\in\mathbb{R}^{n\times{m}}$ between the two graphs $\mathcal{G}_p$ and $\mathcal{G}_{agg}$ above can be defined as 
\begin{equation}
\begin{split}
\pi^{*}_{\mathcal{G}} &= {\arg\min}_{\pi_{\mathcal{G}} \in \Pi(\mu, \nu)} \langle \sigma \mathcal{F}(\textbf{\textit{H}}_{p}, \textbf{\textit{H}}_{\textup{agg}}) \\
&\quad + (1-\sigma) \mathcal{E}(\textbf{\textit{A}}_{p}, \textbf{\textit{A}}_{\textup{agg}}) \otimes \pi_{\mathcal{G}}, \pi_{\mathcal{G}} \rangle,
\end{split}
\end{equation}
with the fused Gromov-Wasserstein distance. By tuning the parameter $\sigma$, we can control the deviation of the learned optimal transport plan between nodes and edges. 

Then we use the backbone model (e.g., HGNNs) to obtain the node representation $\textbf{\textit{Z}}_p$ under each meta-path view. The node representation here contains the semantic information of the $p$-th meta-path, and the node representation $\textbf{\textit{Z}}_{\rm{agg}}$ in the aggregated view calculated in Section 4.2. We can use Equation (8) directly to calculate the optimal plan $\pi^*_{\mathcal{Z}}\in\mathbb{R}^{n\times{m}}$ as
\begin{equation}
    \pi^*_{\mathcal{Z}}={\arg\min}_{\pi_{\mathcal{Z}}\in\Pi\left(\mu, \nu\right)}\langle{\mathcal{R}\left(\textbf{\textit{Z}}_p, \textbf{\textit{Z}}_{\rm{agg}}\right), \pi_{\mathcal{Z}}}\rangle,
\end{equation}
where $\mathcal{R}\left(\textbf{\textit{Z}}_p, \textbf{\textit{Z}}_{\rm{agg}}\right)_{ij}=\mathcal{C}_Z\left(\textbf{\textit{Z}}_{p}^{i}, \textbf{\textit{Z}}_{\rm{agg}}^{j}\right)$ , $\textbf{\textit{Z}}_p$ and $\textbf{\textit{Z}}_{\rm{agg}}$ denote the node representations corresponding to $\mathcal{G}_p$ and $\mathcal{G}_{\rm{agg}}$, respectively.

\textbf{Optimal Transport Alignment.} Obviously, if the node embedding learned by the encoder can capture the matching information with other graphs while preserving its own information, it may obtain better node representations. Therefore, we force the encoder to preserve the matching relationship in graph space by aligning the optimal transfer plan between two graphs with the optimal transport plan between their corresponding node embeddings. We minimize the difference between the optimal transport plans in the two spaces as the alignment loss as follows
\begin{equation}
    \mathcal{L}_{\rm{mat}}=\Theta\left(\pi^*_{\mathcal{G}}, \pi^*_{\mathcal{Z}}\right),
\end{equation}
where $\Theta\left(\cdot, \cdot\right)$ denotes the discrepancy function, it can be any commonly used metric. For example, we can choose the Frobenius-norm as $\Theta\left(\cdot, \cdot\right)={\Vert{\cdot - \cdot}\Vert}_F$

In addition, to guide the encoder to retain the structural information of the original image while learning node embedding, we correct the cost matrix $\mathcal{R}\left(\textbf{\textit{Z}}_p, \textbf{\textit{Z}}_{\rm{agg}}\right)$ between representations, which implies the implicit structure relationships between nodes, in the representation space as follow
\begin{equation}
\begin{split}
     \mathcal{L}_{\rm{str}}=\Theta(\sigma\mathcal{F}\left(\textbf{\textit{H}}_p, \textbf{\textit{H}}_{\rm{agg}}\right)+\left(1-\sigma\right)\mathcal{E}\left(\textbf{\textit{A}}_p, \textbf{\textit{A}}_{\rm{agg}}\right)\otimes\\\pi^*_{\mathcal{G}}, \mathcal{R}\left(\textbf{\textit{Z}}_p,\textbf{\textit{ Z}}_{\rm{agg}}\right)).
\end{split}
\end{equation}
To this end, we define the overall heterogeneous graph transport alignment loss as
\begin{equation}
    \mathcal{L}=\mathcal{L}_{\rm{mat}}+\rho\mathcal{L}_{\rm{str}},
\end{equation}
where $\rho$ is the trade-off parameter.

\subsection{Comparison with Graph Contrastive Learning}

A common way of traditional contrastive objective is contrasting views at the node level. For the representations (positive views) $h_i^1$ and $h_i^2$ of same node $i$ in two augmented graphs $\mathcal{G}_1$ and $\mathcal{G}_2$, the pairwise contrastive loss can be defined as
\begin{equation}
    \mathcal{L}_{\rm{com}}= \sum_i{\rm{log}\left({\cfrac{e^{{\mathcal{I}\left(\textbf{h}_\textit{i}^1, \textbf{h}_\textit{i}^2\right)/\tau}}}{e^{{\mathcal{I}\left(\textbf{h}_\textit{i}^1, \textbf{h}_\textit{i}^2\right)/\tau}}+\sum_{\textit{k}\not=\textit{i}}e^{{\mathcal{I}\left(\textbf{h}_\textit{i}^1,\textit{ }\textbf{h}_\textit{k}^2\right)/\tau}}}}\right)}
\end{equation}
 where $\mathcal{I}$ is a discriminator that maps two views to an agreement (similarity) score, such as the inner product. $\tau$ denotes the temperature.

Although our objective function deviates from the contrastive loss, as outlined in Equation (20), we identify a striking resemblance in the foundational algorithmic principles governing both methodologies. In this examination, we concentrate on the node level. When the attributes and contextual information of a given node $v_i$, demonstrate similarity across diverse semantic perspectives, the associated cost for these views diminishes, nearly reaching zero. This reduction in cost leads to an optimal transport strategy that favors a high probability of aligning the representations of these nodes. Leveraging Equations (17) and (18), the HGOT model adeptly fine-tunes both the matching probabilities and costs linked to the corresponding node representations within the output domain. Consequently, this ensures that the representations of two analogous nodes are sufficiently congruent. In contrast, for two distinct nodes $v_i$ and $v_k$, the costs linked to their respective node views across varying semantic viewpoints are relatively higher. This discrepancy necessitates an adjustment in the opposite direction, ensuring that the representations of two dissimilar nodes are sufficiently differentiated.

From the perspective of contrastive learning, the process described above is in harmony with its core objectives. Nevertheless, there is a notable disparity between graph transport alignment and graph contrastive learning. HGOT capitalizes on calibration signals sourced directly from the graph space. This unique calibration signal empowers us to circumvent the need to distinguish between positive and negative samples.

\section{Experiments}
\begin{figure*}[!h]
    \centering
   \textit{ Table 1}. Experimental results on node classification.
    \includegraphics[width=1\linewidth]{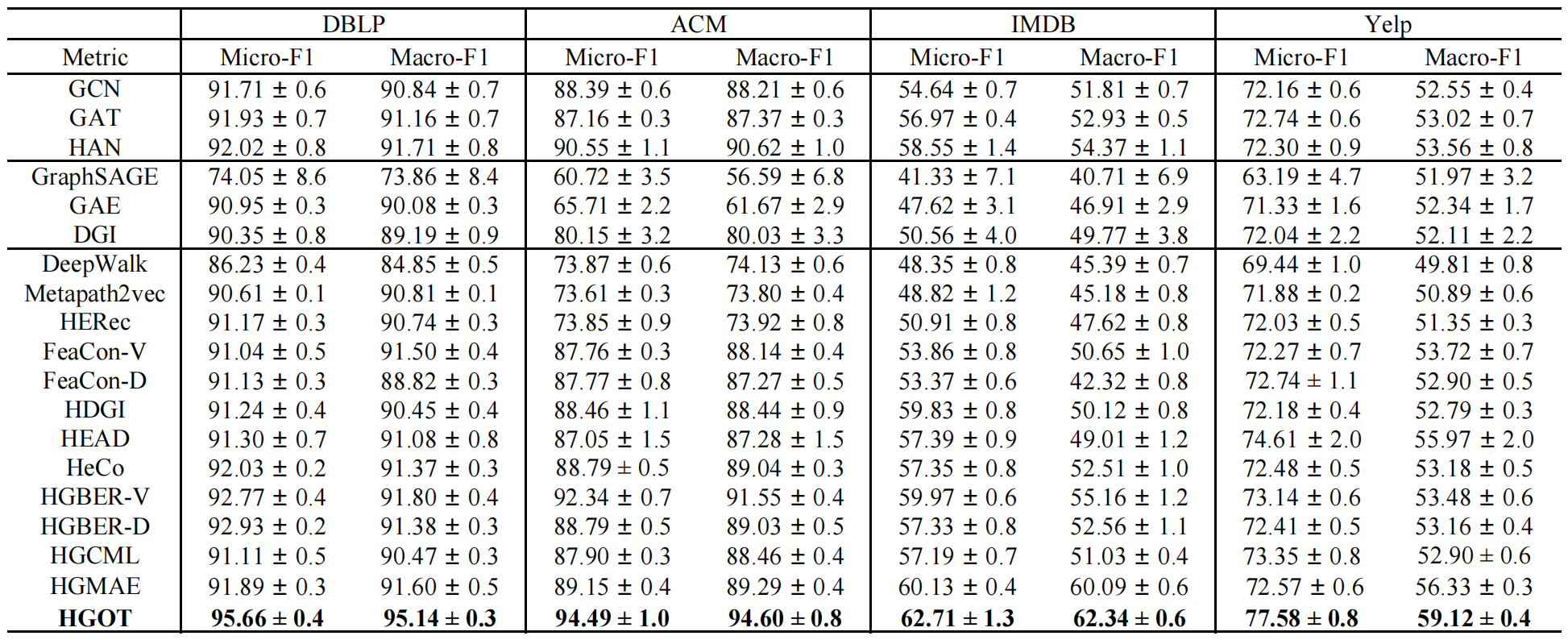}

\end{figure*}

In this section, we conduct extensive experiments to verify the effectiveness of the proposed HGOT. Specifically, we first demonstrate the advantages of our proposed model over other methods in different tasks, including node classification, node clustering, and visualization of node representations. 
Then, we conduct other experiments to verify the effectiveness of the modules in our method, including the ablation study and parameter sensitivity analysis.

\subsection{Experimental Setup}

\textbf{Datasets.} Four public datasets (DBLP, ACM, IMDB and Yelp) are used to demonstrate the effectiveness of the proposed method. The detailed information on datasets is provided in attached page.

\textbf{Baselines.} We compare the proposed HGOT with three categories of baselines: \textbf{supervised methods:} GCN (\textcolor[RGB]{25,25,112}{T. N. Kipf et al. 2016}), GAT (\textcolor[RGB]{25,25,112}{P. Veliˇckovi´c et al. 2017}), a \textbf{semi-supervised heterogeneous method:} HAN (\textcolor[RGB]{25,25,112}{Wang et al. 2019}), and \textbf{unsupervised methods:} GraphSAGE (\textcolor[RGB]{25,25,112}{William L. et al. 2017}), GAE (\textcolor[RGB]{25,25,112}{T. N. Kipf et al. 2016}), DGI (\textcolor[RGB]{25,25,112}{Petar Velickovic et al. 2019}), DeepWalk (\textcolor[RGB]{25,25,112}{B. Perozzi et al. 2014}), Metapath2vec (\textcolor[RGB]{25,25,112}{Dong et al. 2017}), HERec (\textcolor[RGB]{25,25,112}{Shi et al. 2019}), FeaCon-V, FeaCon-D, HDGI (\textcolor[RGB]{25,25,112}{Ren et al. 2019}), HEAD (\textcolor[RGB]{25,25,112}{Wang et al. 2021}), HeCo (\textcolor[RGB]{25,25,112}{Wang et al, 2021a}), HGBER (\textcolor[RGB]{25,25,112}{Liu et al. 2023}), HGCML  (\textcolor[RGB]{25,25,112}{Wang et al. 2023}) and HGMAE (\textcolor[RGB]{25,25,112}{Tian et al. 2023}).

\textbf{Implementation Details.} For baselines, we adhere to the settings described in their original papers and follow the setup in HeCo. For the proposed HGOT, we use HAN as the backbone model to obtain the node representation $\textbf{\textit{Z}}_p$ under each meta-path view. We search the learning rate from 1e-4 to 5e-3, tune the patience for early stopping from 5 to 20. The
Adam optimizer (\textcolor[RGB]{25,25,112}{Diederik P. Kingma et al. 2015}) is adopted for gradient descent. For all methods, we set the embedding dimension as 64 and report the mean and standard deviation of 10 runs with different random seeds. Among all the methods, the HGBER model has two variants (HGBER-V and HGBER-D) and we compare them both.

\subsection{\textbf{Node Classification}} 

We evaluate different models for the node classification task and report their performances in Table 1. We use Micro-F1 and Macro-F1 as evaluation metrics. Results demonstrate the effectiveness of HGOT as it achieves the best performance over all these baselines. Compared with the homogeneous graph model, HGOT can capture the high-order semantic relationship between nodes to obtain a better node representation. Therefore, our method has a significant improvement in the accuracy of node classification tasks compared to homogeneous graph methods, and even improves by about 34\% compared to GraphSAGE on the ACM dataset. Although methods such as HAN and Metapath2vec have utilized meta-paths, and HeCo has attempted a contrastive learning scheme between meta-paths and network modes, these methods do not study the matching information between meta-path view and aggregated view. Because our model HGOT attempts to learn the matching information from the graph space and representation space, the quality of node representation is improved. Thus our HGOT achieves the best classification performance.

\begin{figure*}[!h]
    \centering
   \textit{ Table 2}. Experimental results on node clustering.
    \includegraphics[width=1\linewidth]{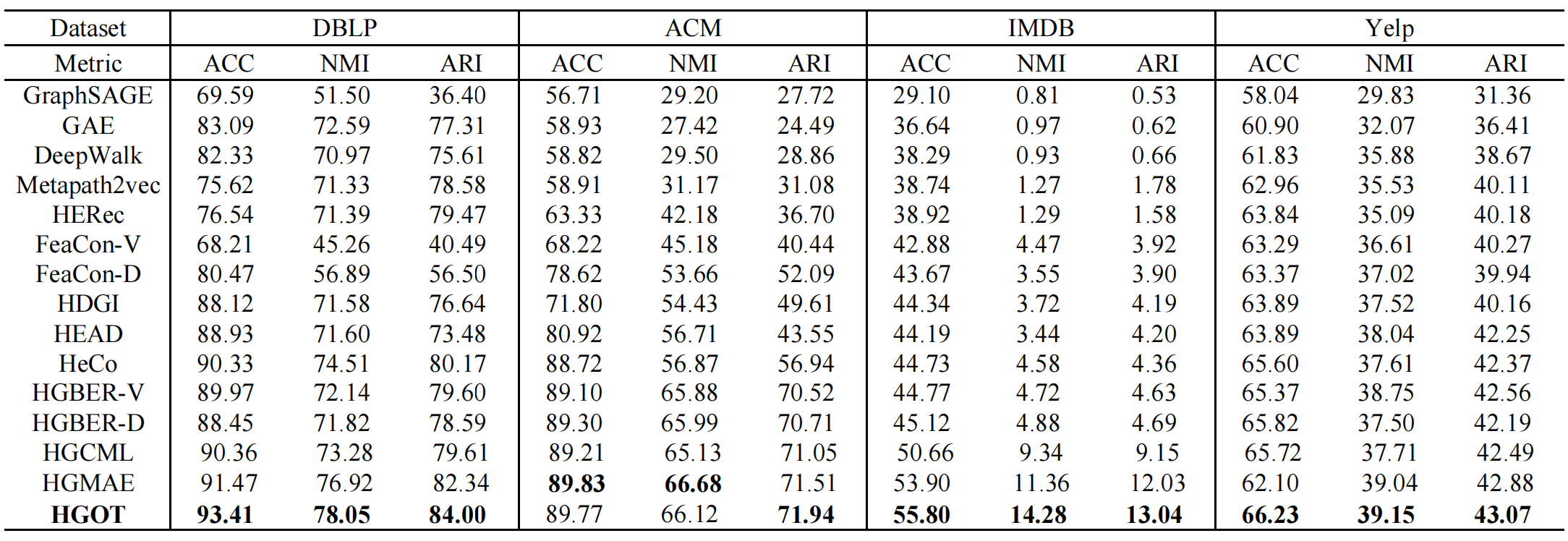}
\end{figure*}

\subsection{\textbf{Node Clustering}}
In this task, we utilize hierarchical clustering algorithm to the learned embeddings of all nodes and the results of the node classification are shown in Table 2. Hierarchical clustering can generate hierarchical clustering structures. Compared with the K-means clustering algorithm, it does not require the number of clusters to be specified in advance. The appropriate number of clusters can be selected as needed, and it can process various types of data. Indicators such as clustering accuracy (ACC), normalized mutual information (NMI), and adjusted rand index (ARI) (\textcolor[RGB]{25,25,112}{Xia et al. 2014}) are used as metrics to assess the quality of the clustering results. It can be seen that HGOT achieves the best performances on most datasets in node clustering task. Notably, HGOT demonstrates an average performance improvement of approximately 2.5\% on the DBLP, IMDB, and Yelp datasets. 

\subsection{\textbf{Visualization}}
\begin{figure}[!t]
    \centering
    \includegraphics[width=1\linewidth]{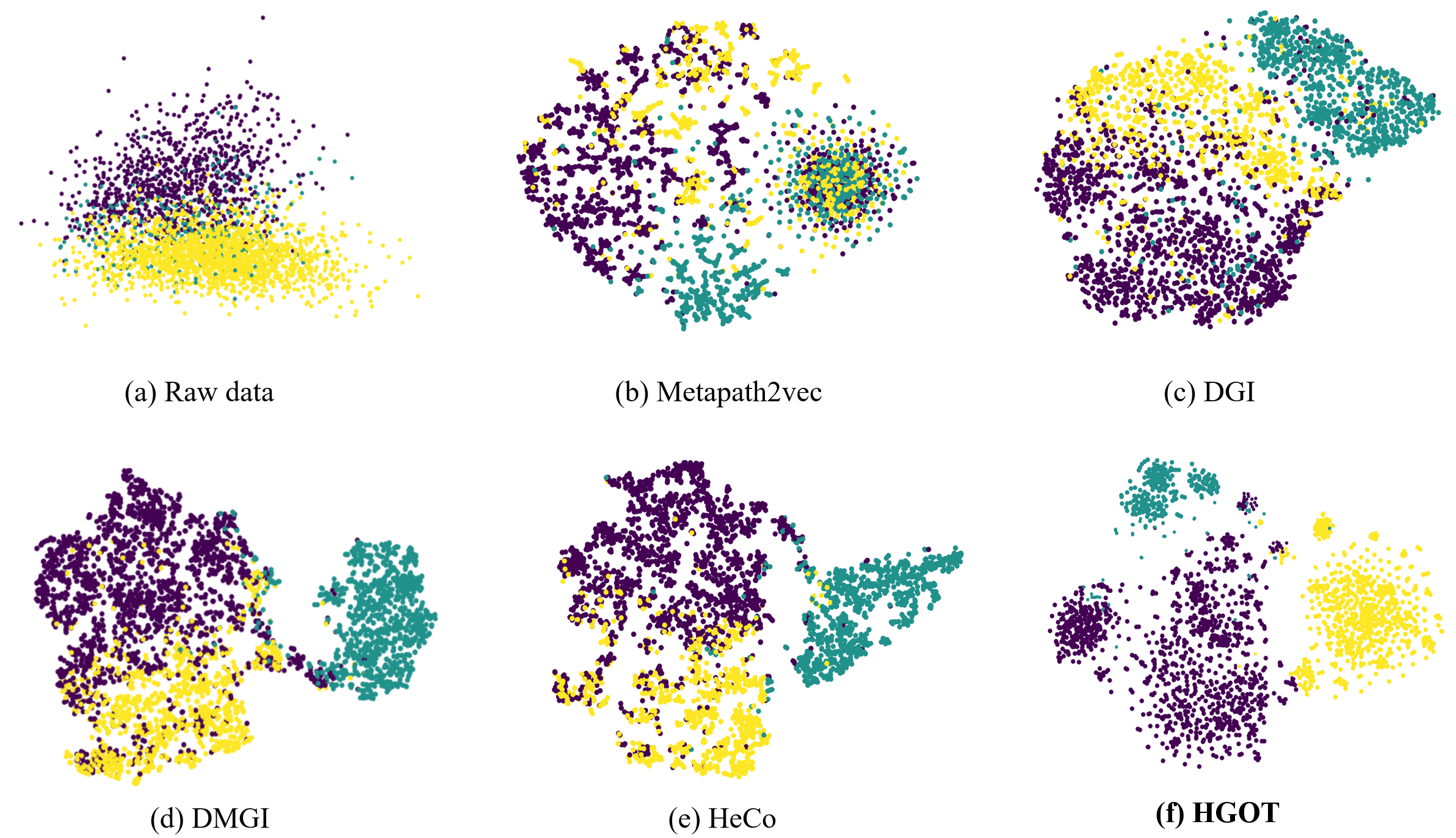}

   \caption{Visualization results of different methods on ACM.}   
\end{figure}
In order to obtain more intuitive evaluation results, we conduct embedding visualization on ACM dataset. We use the t-SNE (\textcolor[RGB]{25,25,112}{Van Der Maaten L. 2014}) dimensionality reduction method to experiment on different models and the results are shown in Figure 2, in which different colors mean different labels. As we can see, the Metapath2vec and DGI method does not classify nodes well, the visualization results show that they cannot make the boundaries between different types of nodes clearer because they cannot fuse all kinds of semantics. Although DMGI and HeCo have achieved the ability to clearly distinguish between various types of nodes, they are still mixed to some degree. The proposed HGOT correctly separates different nodes with relatively clear boundaries, demonstrating the effectiveness of HGOT.

\subsection{\textbf{Ablation Study}}

In this section, we compare HGOT with its variants on  four datasets to verify the effectiveness of three components. HGOT-w/o\#1 does not have the ‘‘aggregated graph generation’’ module. We directly match the two different meta-path views without calculating any information in the aggregated view. 
\begin{figure}[!t]
    \centering
    \includegraphics[width=1\linewidth]{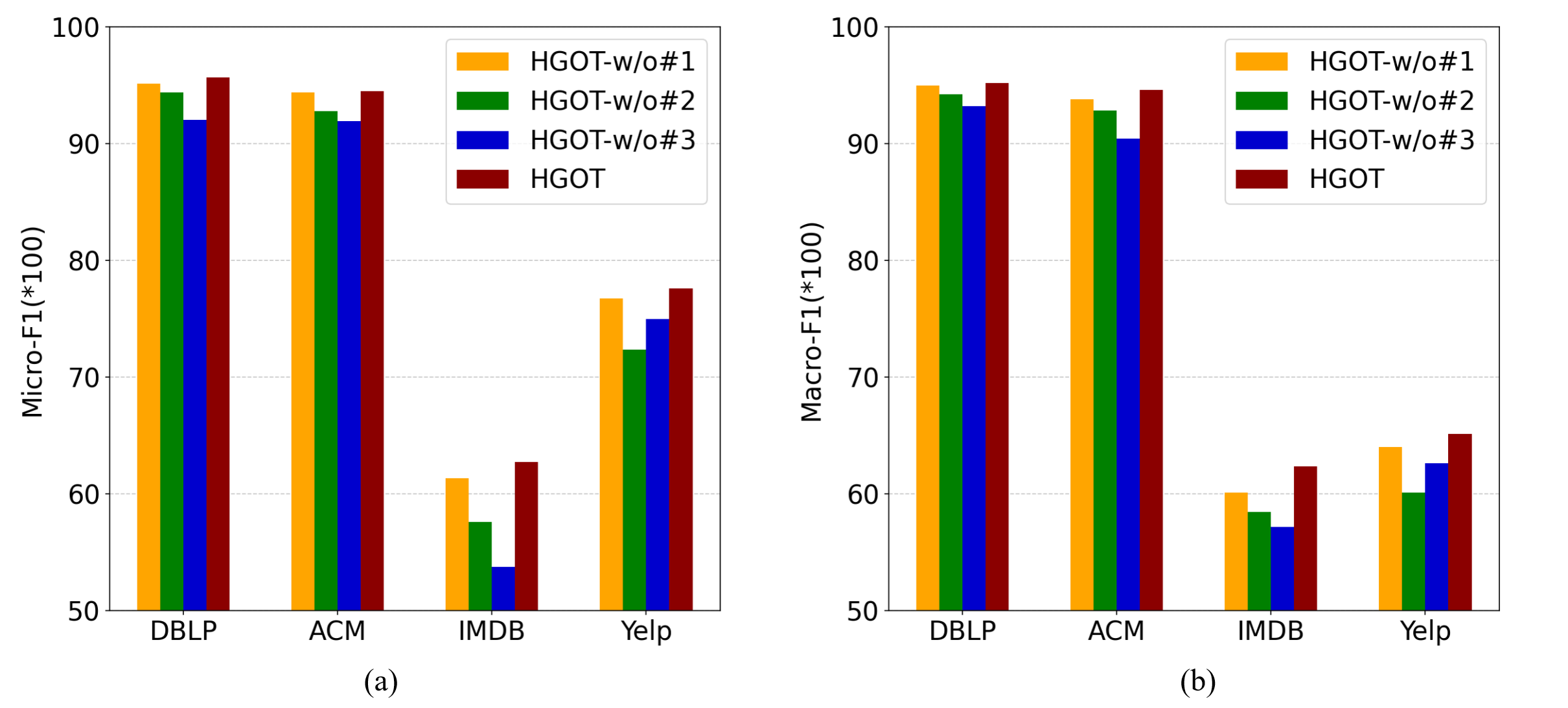}

\caption{Ablation experimental results on four datasets.}
\end{figure}
HGOT-w/o\#2 represents HGOT without implicit structure loss $\mathcal{L}_{\rm{str}}$. HGOT-w/o\#3 drops the alignment of the optimal transport plan, but instead optimize the transport distance directly. In order to make the optimal transport distance between the graph space and the representation space consistent, we replace the matching loss with the following:

\begin{equation}
    \mathcal{L'}_{\rm{mat}}=\vert \mathcal{D}_g\left(\mathcal{G}_p, \mathcal{G}_{\rm{agg}}\right)-\mathcal{D}_n\left(\textbf{\textit{Z}}_p, \textbf{\textit{Z}}_{\rm{agg}}\right)\vert,
\end{equation}

where $\vert\cdot\vert$ denotes absolute value. The results are shown in Figure 3.

\begin{figure}[!h]
    \centering
    \includegraphics[width=1\linewidth]{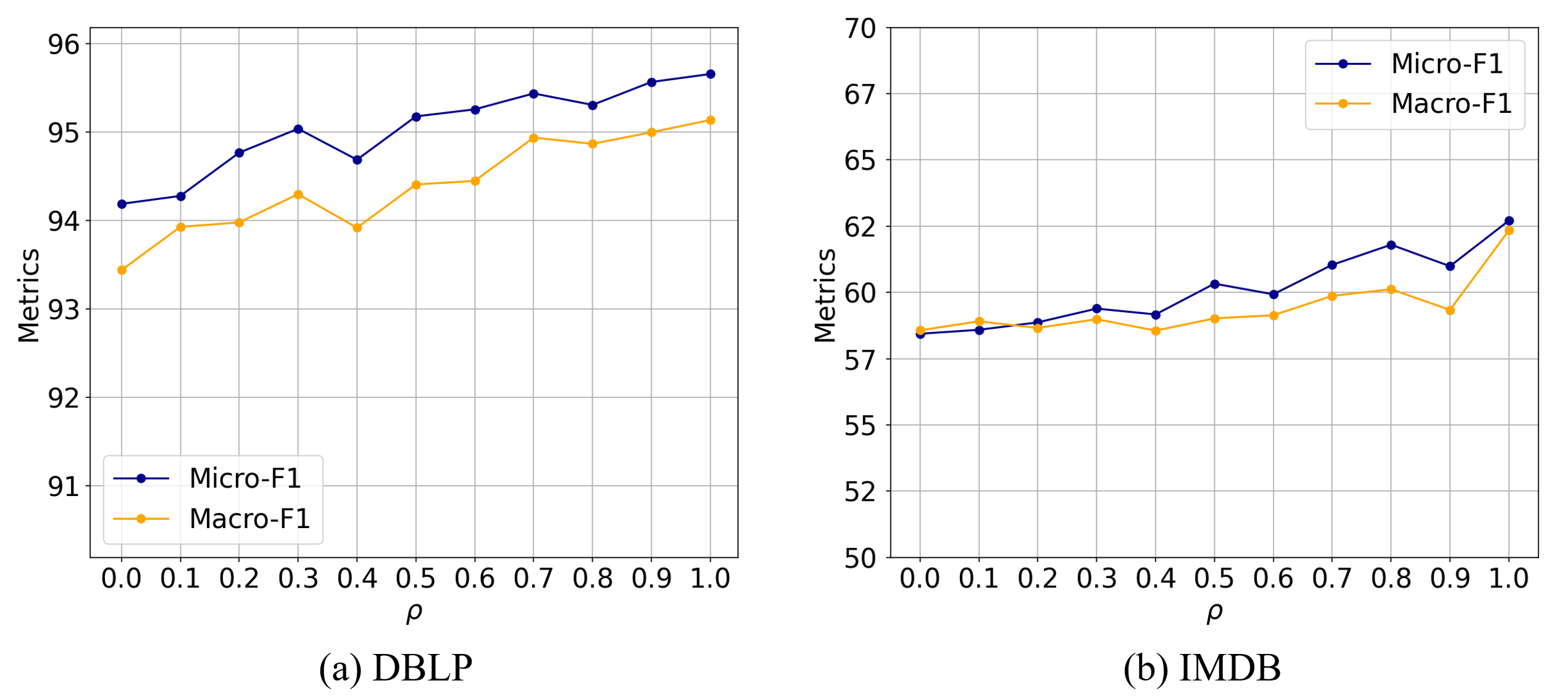}
    \includegraphics[width=1\linewidth]{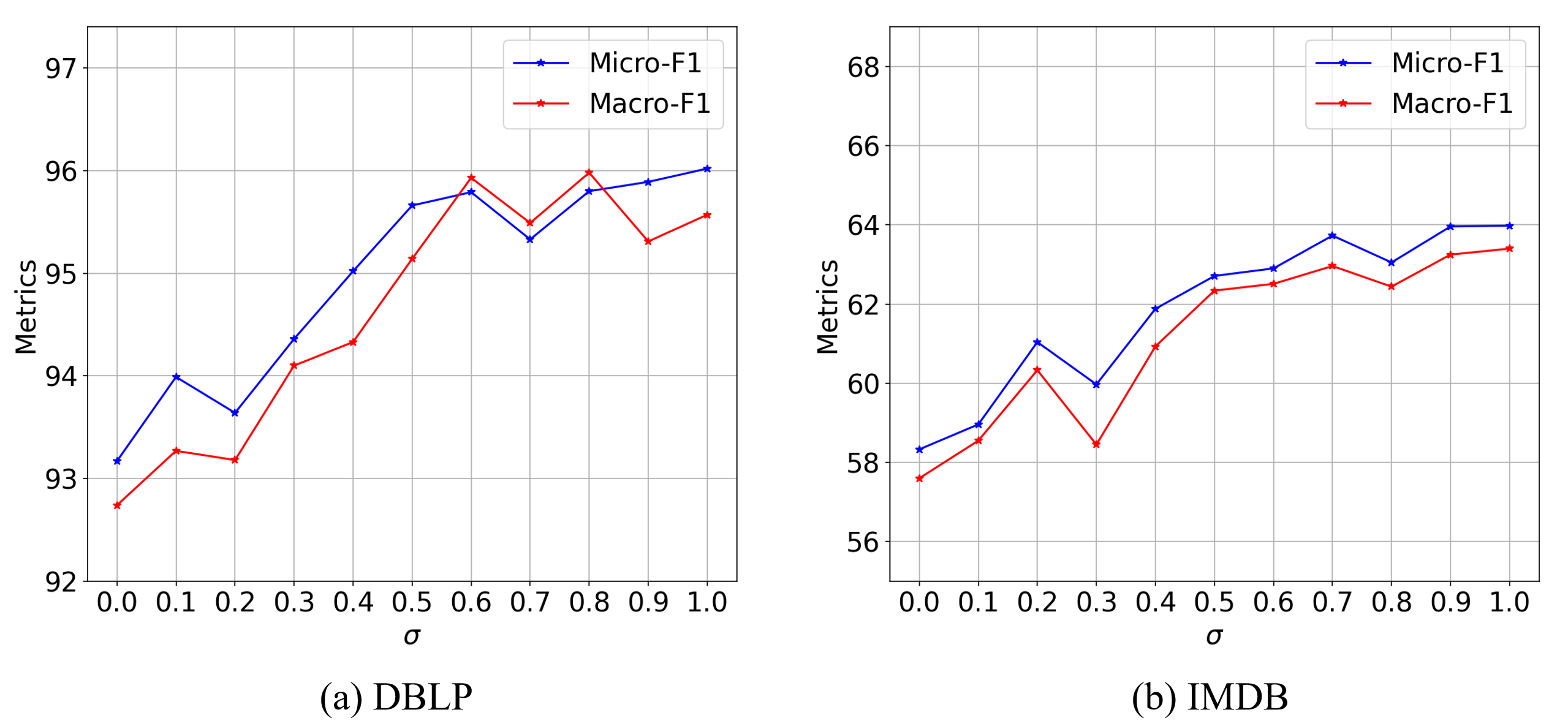}

    \caption{Analysis of hyperparametes $\rho$ and $\sigma$ on DBLP and IMDB datasets.}      
\end{figure}
From Figure 3, conclusions can be drawn as follows: (1) The results of HGOT are consistently better than three variants, indicating the effectiveness of the heterogeneous semantic information aggregation, the implicit structure loss and the alignment of the optimal transport plan strategy. (2) HGOT-w/o\#3 is the worst one, which makes us realize the necessity of the optimal transport plan alignment strategy. (3) The performance of HGOT-w/o\#1 is also very competitive, which demonstrates that the strategy of aggregating all semantic information makes relatively little contribution to the overall model.

\subsection{Parameter Sensitivity}
In this section, we investigate the sensitivity of hyperparameters $\rho$ and $\sigma$ on two datasets. The experimental results are shown in Figure 4. We firstly fix the value of parameter $\sigma$ to 0.5 and analyze the effect of parameter $\rho$ on the prediction of performance.  We can see that as the parameter $\rho$ increases, HGOT shows an overall growth trend on both datasets. HGOT achieves optimal performance when $\rho=1$. However, without the implicit structure loss, HGOT performs not well, which is also shown in the ablation study section. 
In addition, we fix the value of parameter $\rho$ to 1, and analyze the effect of parameter $\sigma$ on the prediction of performance. When $\sigma=0$, HGOT only considers the matching information between edge structures, while when $\sigma=1$, it only considers the matching information between nodes. Figure 4 reports the surprising outcome: With only node attributes for the calculation of the plan, the model achieves outstanding performance on both datasets. However, with both nodes and edges, the result is not optimal. At the same time, our model HGOT performs not well when only edges are considered. These experimental results give us an inspiration: In heterogeneous information networks, the number of edges may far exceed the number of nodes, so we can appropriately abandon some edge information. In this way, performance can be optimized while reducing the time complexity of the model.

\subsection{Complexity Analysis}
We select seven heterogeneous graph self-supervised learning methods and evaluate running time per training epoch on the DBLP dataset. According to the experimental result in Figure 5, we can see that HGOT has the best Micro-F1 score and requires the least training time, because it abandons the complex graph enhancement methods and the positive and negative sample selection process.
The time complexity of the HGOT model mainly hinges on the optimization methods used for Equations (15) and (16). For Equation (15), which includes the fused Gromov-Wasserstein term, we use a conditional gradient (CG) solver for optimization. At each step of this solver, we need to calculate a gradient, and the time it takes for this calculation grows roughly in a cubic way as the number of nodes in the graph increases. On the other hand, Equation (16), which contains the Wasserstein term, can be optimized using the Sinkhorn-Knopp algorithm. This algorithm is quite efficient because its time complexity grows almost quadratically with the graph size. 

\begin{figure}
    \centering
    \includegraphics[width=1\linewidth]{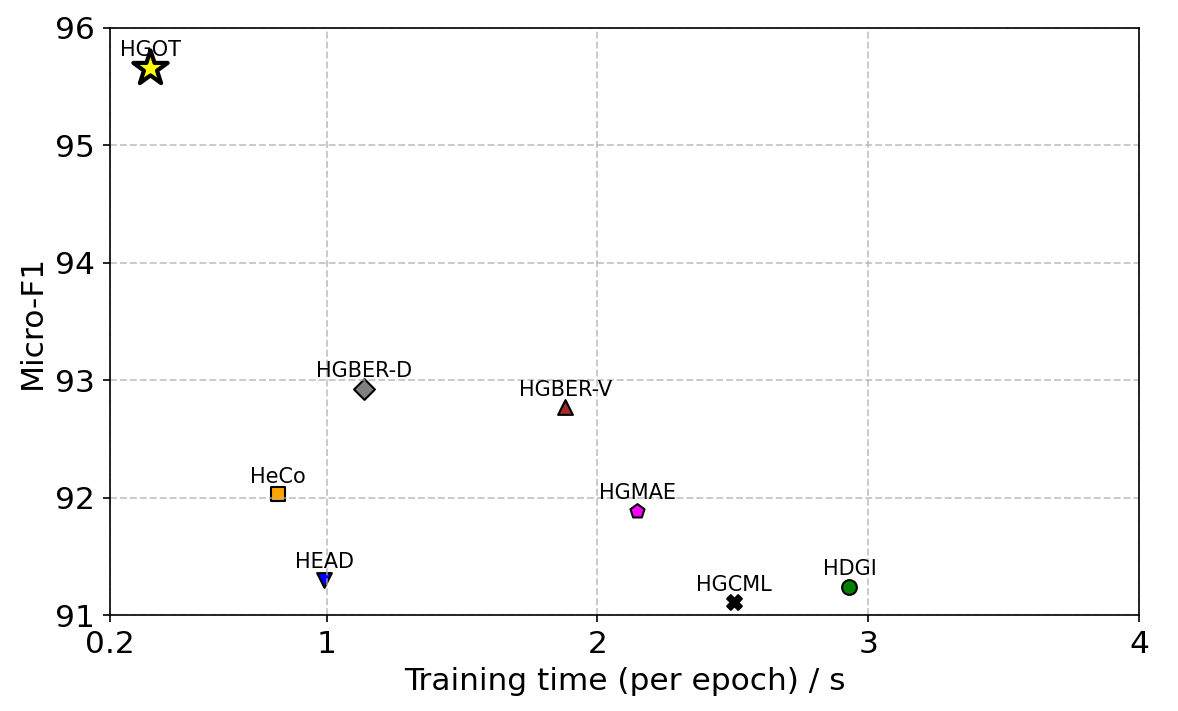}

    \caption{Micro-F1 scores and time consumption on DBLP dataset.}      
\end{figure}
\section{Conclusion}

In this paper, we propose a novel self-supervised heterogeneous graph neural networks with optimal transport, named HGOT. The attention mechanism is employed to obtain an aggregated view which can integrate the semantic information from different meta-paths. Then, HGOT exploits the optimal transport theory to discover the optimal transport plan between the meta-path view and the aggregated view. By aligning the transport plans between graph space and representation sapce, HGOT enforces the backbone model to learn node representations that precisely preserve the matching relationships. Extensive experiments conducted on multiple datasets demonstrate the state-of-the-art performance of the proposed HGOT.

\section*{Acknowledgements}

This work is supported in part by National Natural Science Foundation of China (No. 62371340, 62172052, 62322203), Tianjin Natural Science Foundation Project (No. 24JCZDJC00820, 23JCYBJC00520), Tianjin Measurement Science and Technology Project (NO.2024TJMT061).

\section*{Impact Statement}

This work provides a novel self-supervised learning approach for heterogeneous graphs in real-world scenario, and it aims to promote the development of heterogeneous graph neural networks in machine learning. There are many potential societal consequences of our work, none which we feel must be specifically highlighted here.
\section*{\textbf{References}}

\noindent Wang, X.; Ji, H.; Shi, C.; Wang, B.; Ye, Y.; Cui, P.; and Yu, 
\hangafter=1 \hangindent=10pt
P. S. 2019. Heterogeneous Graph Attention Network. In \textit{WWW}

\noindent Bai, Y.; Ying, Z.; Ren, H.; and Leskovec, J. 2021. Modeling
\hangafter=1 \hangindent=10pt
heterogeneous hierarchies with relation-specifc hyperbolic cones. In \textit{NeurIPS}.

\noindent Cao, Y.; Peng, H.; Wu, J.; Dou, Y.; Li, J.; and Yu, P. S. 2021.
\hangafter=1 \hangindent=10pt
Knowledge-preserving incremental social event detection via heterogeneous gnns. In \textit{WWW}.

\noindent Wang, H.; and Leskovec, J. 2020. Unifying graph convolu
\hangafter=1 \hangindent=10pt
tional neural networks and label propagation. \textit{arXiv preprint }\textit{arXiv:2002.06755}.

\noindent Dong, Y.; Hu, Z.; Wang, K.; Sun, Y.; and Tang, J. 2020. 
\hangafter=1 \hangindent=10pt
Heterogeneous Network Representation Learning. In \textit{IJCAI}.

\noindent Tian, Y.; Zhang, C.; Guo, Z.; Huang, C.; Metoyer, R.; and Chawla, N. 
\hangafter=1 \hangindent=10pt
V. 2022a. RecipeRec: A Heterogeneous Graph Learning Model for Recipe Recommendation. In \textit{IJCAI}.

\noindent Bansal, T.; Juan, D.-C.; Ravi, S.; and McCallum, A. 2019. A2N: 
\hangafter=1 \hangindent=10pt
Attending to neighbors for knowledge graph inference. In \textit{Proceedings of the 57th annual meeting of the as}\textit{sociation for computational linguistics}, 4387–4392.

\noindent Wang, S.; Wei, X.; Nogueira dos Santos, C. N.; Wang, Z.; Nallapati, 
\hangafter=1 \hangindent=10pt
R.; Arnold, A.; Xiang, B.; Yu, P. S.; and Cruz, I. F. 2021a. Mixed-curvature multi-relational graph neural network for knowledge graph completion. In \textit{WWW}, 1761–1771.

\noindent Sun Y, Zhu D, Wang Y, et al. GTC: GNN-transformer co-contrastive 
\hangafter=1 \hangindent=10pt
learning for self-supervised heterogeneous graph representation. Neural Networks, 2025, 181: 106-115. 

\noindent Ji Z, Kong D, Yang Y, et al. ASSL-HGAT: Active semi-supervised 
\hangafter=1 \hangindent=10pt
learning empowered heterogeneous graph attention network. Knowledge-Based Systems, 2024, 290: 156-167. 

\noindent Weihua Hu, Bowen Liu, Joseph Gomes, Marinka Zitnik, Percy 
\hangafter=1 \hangindent=10pt
Liang, Vijay S. Pande, and Jure Leskovec. 2020. Strategies for Pre-training Graph Neural Networks. In \textit{ICLR}.

\noindent Wang, X.; Liu, N.; Han, H.; and Shi, C. 2021a. Self-supervised 
\hangafter=1 \hangindent=10pt
heterogeneous graph neural network with co-contrastive learning. In \textit{KDD}.

\noindent Zhao Z, Zhu Z, Liu Y, et al. Heterogeneous Graph Contrastive 
\hangafter=1 \hangindent=10pt
Learning with Augmentation Graph. IEEE Transactions on Artificial Intelligence, 2024.

\noindent Zhao M, Jia A L. Dahgn: Degree-aware heterogeneous graph 
\hangafter=1 \hangindent=10pt
neural network. Knowledge-Based Systems, 2024, 285: 135-145. 

\noindent Xie Y, Yu C, Jin X, et al. Heterogeneous graph contrastive 
\hangafter=1 \hangindent=10pt
learning for cold start cross-domain recommendation. Knowledge-Based Systems, 2024: 2054-2065. 

\noindent Liu Y, Xia L, Huang C. 2024. Selfgnn: Self-supervised graph neural 
\hangafter=1 \hangindent=10pt
networks for sequential recommendation. In \textit{SIGIR}. 

\noindent Chen L, Cao J, Liang W, et al. Geography-aware Heterogeneous \hangafter=1 \hangindent=10pt 
Graph Contrastive Learning for Travel Recommendation. ACM Transactions on Spatial Algorithms and Systems, 2024. 

\noindent Yang, C.; Xiao, Y.; Zhang, Y.; Sun, Y.; and Han, J. \hangafter=1 \hangindent=10pt 
Heterogeneous Network Representation Learning: A Unifed Framework with Survey and Benchmark. IEEE Transactions on Knowledge and Data Engineering, 2020, 34(10): 4854-4873.

\noindent Wang, X.; Bo, D.; Shi, C.; Fan, S.; Ye, Y.; and Philip, S. Y. A 
\hangafter=1 \hangindent=10pt 
survey on heterogeneous graph embedding: methods, techniques, applications and sources. IEEE Transactions on Big Data, 2022, 9(2): 415-436.

\noindent Fan, Y.; Ju, M.; Zhang, C.; and Ye, Y. 2022. Heterogeneous 
\hangafter=1 \hangindent=10pt 
temporal graph neural network. In \textit{SDM}.

\noindent Fu, X.; Zhang, J.; Meng, Z.; and King, I. 2020. Magnn: Metapath 
\hangafter=1 \hangindent=10pt 
aggregated graph neural network for heterogeneous graph embedding. In \textit{WWW}.

\noindent Zhang Q, Zhao Z, Zhou H, et al. Self-supervised contrastive 
\hangafter=1 \hangindent=10pt 
learning on heterogeneous graphs with mutual constraints of structure and feature. Information Sciences, 2023, 640: 119026. 

\noindent Kipf, T. N.; and Welling, M. 2016. Variational graph 
\hangafter=1 \hangindent=10pt 
auto-encoders. \textit{arXiv preprint arXiv:1611.07308}.

\noindent Park, J.; Lee, M.; Chang, H. J.; Lee, K.; and Choi, J. Y. 2019. 
\hangafter=1 \hangindent=10pt 
Symmetric graph convolutional autoencoder for unsupervised graph representation learning. In \textit{ICCV}.

\noindent Wang Z, Li Q, Yu D, et al. 2023. Heterogeneous graph contrastive 
\hangafter=1 \hangindent=10pt 
multi-view learning. In SDM.

\noindent Tian Y, Dong K, Zhang C, et al. 2023. Heterogeneous graph masked 
\hangafter=1 \hangindent=10pt 
autoencoders. In \textit{AAAI}. 

\noindent Cédric Villani et al. \textit{Optimal transport: old and new}, volume 338. 
\hangafter=1 \hangindent=10pt 
Springer, 2009.

\noindent Nicolas Bonneel, Michiel Van De Panne, Sylvain Paris, and 
\hangafter=1 \hangindent=10pt 
Wolfgang Heidrich. Displacement interpolation using lagrangian mass transport. In \textit{Proceedings of the 2011 SIGGRAPH Asia }\textit{Conference}, pages 1–12, 2011.

\noindent Jiezhang Cao, Langyuan Mo, Yifan Zhang, Kui Jia, Chunhua Shen.2019. \hangafter=1 \hangindent=10pt 
and Mingkui Tan. Multi-marginal wasserstein gan. In \textit{NeurIPS}.

\noindent Xiang Gu, Yucheng Yang, Wei Zeng, Jian Sun, and Zongben Xu. 2022. \hangafter=1 \hangindent=10pt 
Keypoint-guided optimal transport with applications in heterogeneous domain adaptation. In \textit{NeurIPS}.

\noindent Facundo Mémoli. Gromov–wasserstein distances and the metric \hangafter=1 \hangindent=10pt 
approach to object matching. Foundations of Computational Mathematics, 2011, 11:417–487.

\noindent Vayer Titouan, Nicolas Courty, Romain Tavenard, and Rémi. 2019. \hangafter=1 \hangindent=10pt 
Flamary. Optimal transport for structured data with application on graphs. In \textit{ICML}.

\noindent Cédric Vincent-Cuaz, Rémi Flamary, Marco Corneli, Titouan. 2022. \hangafter=1 \hangindent=10pt 
Vayer, and Nicolas Courty. Semi-relaxed gromov-wasserstein divergence and applications on graphs. In \textit{ICLR}.

\noindent Hongteng Xu, Dixin Luo, and Lawrence Carin. 2019. \hangafter=1 \hangindent=10pt Scalable gromov-wasserstein learning for graph partitioning and matching. In \textit{NeurIPS}.

\noindent Wang Y, Zhao Y, Wang D Z, et al. 2024. GALOPA: graph transport 
\hangafter=1 \hangindent=10pt
learning with optimal plan alignment. In \textit{NeurIPS}. 

\noindent T. N. Kipf and M. Welling. Semi-supervised classification with \hangafter=1 \hangindent=10pt
graph convolutional networks, 2016, \textit{arXiv:1609.02907}.

\noindent P. Veliˇckovi´c, G. Cucurull, A. Casanova, A. Romero, P. Liò, and Y. \hangafter=1 \hangindent=10pt
Bengio. Graph attention networks, 2017, \textit{arXiv:1710.10903}.

\noindent William L. Hamilton, Zhitao Ying, and Jure Leskovec. 2017. \hangafter=1 \hangindent=10pt
Inductive Representation Learning on Large Graphs. In \textit{NeurIPS}. 

\noindent Thomas N. Kipf and Max Welling. 2016. Variational graph auto-\hangafter=1 \hangindent=10pt
encoders. \textit{arXiv }\textit{preprint arXiv:1611.07308 }(2016).

\noindent Petar Velickovic, William Fedus, William L. Hamilton, Pietro Liò, \hangafter=1 \hangindent=10pt
Yoshua Bengio, and R. Devon Hjelm. 2019. Deep Graph Infomax. In \textit{ICLR}.

\noindent B. Perozzi, R. Al-Rfou, and S. Skiena, 2014. DeepWalk: Online learning \hangafter=1 \hangindent=10pt
of social representations. In \textit{KDD}.

\noindent Y. Dong, N. V. Chawla, and A. Swami, 2017. Metapath2vec: Scalable \hangafter=1 \hangindent=10pt
representation learning for heterogeneous networks. In \textit{KDD}.

\noindent C. Shi, B. Hu, W. X. Zhao, and P. S. Yu. Heterogeneous \hangafter=1 \hangindent=10pt
information network embedding for recommendation. IEEE Trans. Knowl. Data Eng., vol. 31, no. 2, pp. 357–370, 2019.

\noindent Y. Ren, B. Liu, C. Huang, P. Dai, L. Bo, and J. Zhang, \hangafter=1 \hangindent=10pt
Heterogeneous deep graph infomax. 2019, \textit{arXiv:1911.08538}.

\noindent R. Wang, C. Shi, T. Zhao, X. Wang, and Y. F. Ye, Heterogeneous \hangafter=1 \hangindent=10pt
information network embedding with adversarial disentangler. IEEE Trans. Knowl. Data Eng., Jul. 13, 2021.

\noindent Liu Y, Fan L, Wang X, et al. HGBER: Heterogeneous graph neural \hangafter=1 \hangindent=10pt
network with bidirectional encoding representation. IEEE Transactions on Neural Networks and Learning Systems, 2023. 

\noindent Diederik P. Kingma and Jimmy Ba. 2015. Adam: A Method for \hangafter=1 \hangindent=10pt
Stochastic Optimization. In \textit{ICLR}.

\noindent R. Xia, Y. Pan, L. Du, and J. Yin. 2014. Robust multi-view spectral \hangafter=1 \hangindent=10pt
clustering via low-rank and sparse decomposition, In \textit{AAAI}.

\noindent Van Der Maaten L. Accelerating t-SNE using tree-based \hangafter=1 \hangindent=10pt
algorithms. The Journal of Machine Learning Research, 2014, 15(1): 3221-3245.




\newpage
\appendix
\onecolumn
\section{\textbf{Datasets Details}}
The following four public datasets are used to verify the effectiveness of the proposed method.

\textit{\textbf{DBLP:} }DBLP is an integrated database of English-language literature in the field of computer with the results of research as the author as the core. In this article, a subset of DBLP which consists of 14 328 papers (\textit{P}), 4057 authors(\textit{A}), 20 conferences(\textit{C}), and 8789 terms(\textit{T}) is extracted. We divide the authors into four categories according to their research areas: \textit{database, data mining, }\textit{machine learning, information retrieval}. Depending on the conferences linked to the authors, we classify the authors into the appropriate research areas. The attribute information of authors is extracted from keywords. Three meta-paths, APA, APCPA, and APTPA are extracted for use with experiments.

\textit{\textbf{ACM:} }ACM is a paper network contrived from the ACM dataset using papers published in ACM Knowledge Discovery and Data Mining (KDD), SIGMOD, SIGCOMM, Mobi-COMM, and VLDB. ACM dataset consists of 3025 papers (\textit{P}), 5835 authors (\textit{A}), and 56 subjects (\textit{S}). The papers are divided into three categories: \textit{database, wireless communica}\textit{tion, data mining}. Depending on the conferences linked to the papers, we classify the papers into the appropriate categories. The attribute information of papers is extracted from keywords. Two meta-paths PAP, PSP are extracted for use with experiments.

\textit{\textbf{IMDB:}} IMDB is a link dataset built with permission from the Internet Movie Data (IMDB). A subset of IMDB which contains 4780 movies (\textit{M}), 5841 actors (\textit{A}), and 2269 directors (\textit{D}) is extracted. We divide the movies into three categories: \textit{Action, Comedy, Drama }according to their genre. The attribute information of movies is extracted from plots. Two meta-paths \textit{MAM}, \textit{MDM }are extracted for use with experiments.

\textit{\textbf{Yelp:}} The Yelp dataset contains 2614 businesses (\textit{B}), 1286 users (\textit{U}), two reviews (\textit{R}), two services (\textit{S}), and nine rating levels (\textit{L}). The business nodes are labeled by their category. The node features are constructed by the bag-of-words representation of the related keywords. Four meta-paths, BUB, BRB, BSB, and BLB, are extracted for use with experiments.

\section{\textit{Dimension Analysis}}
We further perform dimension sensitivity analysis to show the impact of hidden dimensions in Figure 5. In particular, we search the number of hidden dimensions from \{64, 128, 256, 512, 1024\}, as shown the inflection point of the discount in the figure. It can be seen from the figure, the
optimal hidden dimension can be different across different datasets. As shown in the Figure 6, the best hidden dimension on the DBLP dataset is 256, while on the IMDB is 512. In addition, as the hidden dimension increases, the model effect may be better. For example, when the hidden dimension is only 64 dimensions, the model effect is not good, because using a small hidden dimension could prevent the model from fully capturing the knowledge. However, too wide a hidden dimension could decentralize the model’s focus on meaningful information. Therefore, the model under the standard latent dimension will capture more complex and useful information, and the effect can be optimal. 
\begin{figure}[!h]
    \centering
    \includegraphics[width=0.8\linewidth]{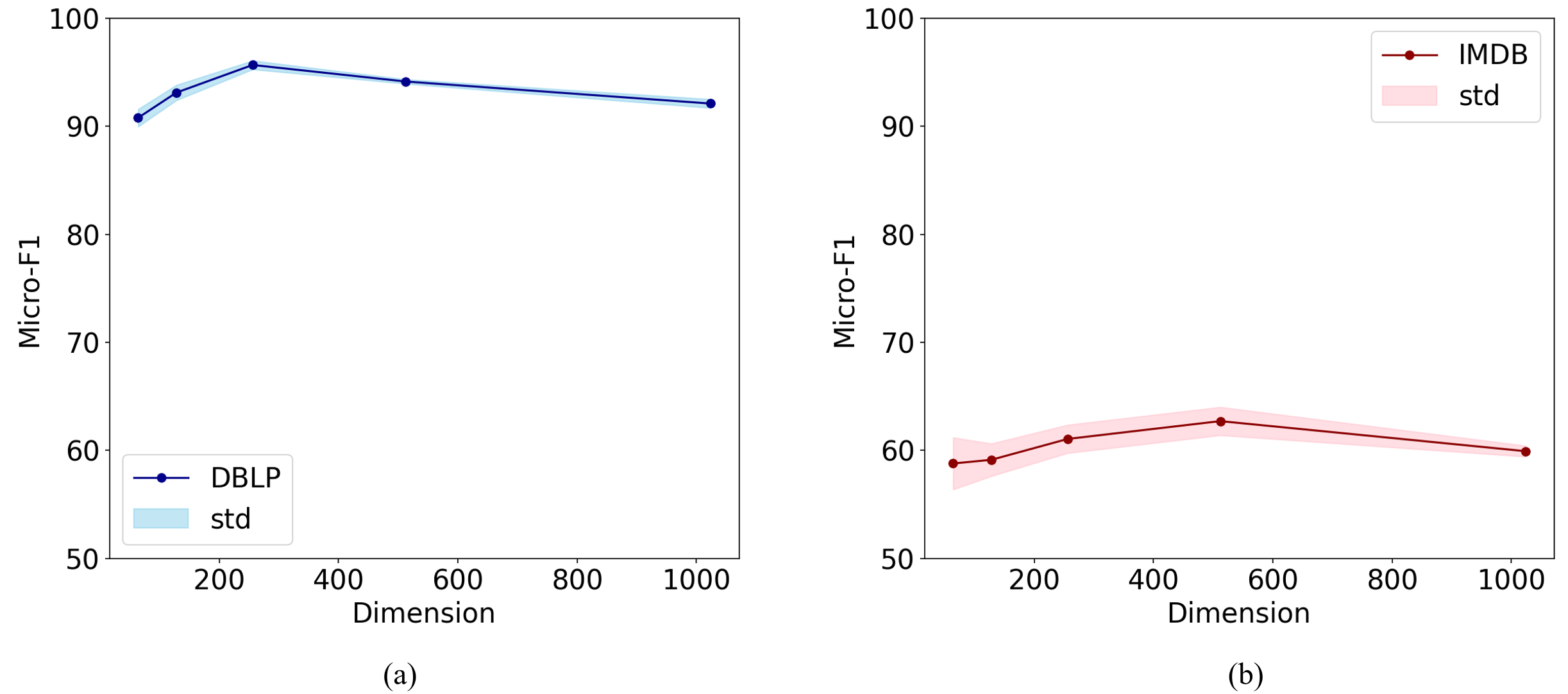}
    \caption{Performance of HGOT under different hidden dimensions.}
    \label{fig:enter-label}
\end{figure}

\section{\textbf{General Settings}}
\textbf{Environment. }The environment in which we run experiments is:

\begin{itemize}
    \item Operating system: Microsoft Windows 11 64-bit
    \item CPU information: 12th Gen Intel(R) Core(TM) i9-12900H (20 CPUs), ~2.5GHz
    \item GPU information: NVIDIA GeForce RTX 3070 Ti Laptop GPU
\end{itemize}

\textbf{Resources. }The address and licenses of all datasets are as follows:

\begin{itemize}
    \item DBLP: https://dblp.uni-trier.de
    \item ACM: http://dl.acm.org/
    \item IMDB: http://komarix.org/ac/ds/
    \item Yelp: https://www.yelp.com/dataset
\end{itemize}


\end{document}